\documentclass[acmtog,prologue,table,xcdraw]{acmart}
\acmSubmissionID{902}

\usepackage{booktabs} 
\usepackage[ruled]{algorithm2e} 

\SetAlFnt{\small}
\SetAlCapFnt{\small}
\SetAlCapNameFnt{\small}
\SetAlCapHSkip{0pt}

\setcopyright{rightsretained}
\acmJournal{TOG}
\acmYear{2023} \acmVolume{42} \acmNumber{6} \acmArticle{204} \acmMonth{12} \acmPrice{15.00}\acmDOI{10.1145/3618401}
\usepackage{graphics}
\usepackage{overpic}
\usepackage{overpic} %
\usepackage{color}
\usepackage{multirow}
\usepackage{paralist}
\usepackage{makecell}
\usepackage{soul} %
\usepackage{pifont}%
\newcommand{\cmark}{\ding{51}}%
\newcommand{\xmark}{\ding{55}}%
\usepackage{tabularx}
\usepackage{svg}
\usepackage{amsmath}
\usepackage{subcaption}
\usepackage{graphicx}
\usepackage[export]{adjustbox}
\usepackage{tikz}
\usetikzlibrary{matrix,calc}
\definecolor{turquoise}{cmyk}{0.65,0,0.1,0.3}
\definecolor{purple}{rgb}{0.65,0,0.65}
\definecolor{dark_green}{rgb}{0, 0.5, 0}
\definecolor{orange}{rgb}{0.8, 0.6, 0.2}
\definecolor{red}{rgb}{0.8, 0.2, 0.2}
\definecolor{darkred}{rgb}{0.6, 0.1, 0.05}
\definecolor{blueish}{rgb}{0.0, 0.3, .6}
\definecolor{light_gray}{rgb}{0.7, 0.7, .7}
\definecolor{pink}{rgb}{1, 0, 1}
\definecolor{greyblue}{rgb}{0.25, 0.25, 1}
\definecolor{forestgreen}{rgb}{0.0, 0.2, 0.13}
\definecolor{darkolivegreen}{rgb}{0.33, 0.42, 0.18}
\newif\ifshowcomments

\newcommand{\moniker}{FLARE}
\newcommand{\fmlp}{\mathcal{M}}
\newcommand{\dmlp}{\mathcal{D}}
\newcommand{\smlp}{\mathcal{S}_{\psi}}
\newcommand{\intmlp}{\mathcal{L}}
\newcommand{\xc}{\mathbf{\mathbf{x_{c}}}}

\newcommand{\nd}{\mathbf{\mathbf{n_{d}}}}
\newcommand{\albedo}{\rho}
\newcommand{\roughness}{r}
\newcommand{\x}{\mathbf{\mathbf{x}}}

\newcommand{\xd}{\mathbf{\mathbf{x_{d}}}}

\newcommand{\flameshape}{\beta}
\newcommand{\flamepose}{\theta}
\newcommand{\flameexpr}{\psi}
\newcommand{\real}{\mathbb{R}}
\newcommand{\wi}{\omega_i}
\newcommand{\wo}{\omega_o}
\newcommand{\refl}{\omega_r}
\newcommand{\n}{\textbf{n}}

\newcommand{\vtx}{\textbf{v}}
\newcommand{\lightmlp}{lighting MLP}
\newcommand{\Lightmlp}{Lighting MLP}

\usetikzlibrary{external}
\begin{document}
\title{FLARE: Fast Learning of Animatable and Relightable Mesh Avatars}
\author{Shrisha Bharadwaj}
\orcid{0009-0000-7390-456X}
\affiliation{%
 \institution{Max Planck Institute for Intelligent Systems}
 \city{Tübingen}
 \country{Germany}}
\email{shrisha.bharadwaj@tuebingen.mpg.de}
\author{Yufeng Zheng}
\orcid{0009-0001-7828-9428}
\affiliation{%
 \institution{ETH Zürich}
 \city{Zürich}
 \country{Switzerland}}
\affiliation{%
 \institution{Max Planck Institute for Intelligent Systems}
 \city{Tübingen}
 \country{Germany}}
\email{yufeng.zheng@inf.ethz.ch}
\author{Otmar Hilliges}
\orcid{0009-0001-7828-9428}
\affiliation{%
 \institution{ETH Zürich}
 \city{Zürich}
 \country{Switzerland}}
\email{otmar.hilliges@inf.ethz.ch}
\author{Michael J.~Black}
\orcid{0000-0001-6077-4540}
\affiliation{%
 \institution{Max Planck Institute for Intelligent Systems}
 \city{Tübingen}
 \country{Germany}}
\email{black@tuebingen.mpg.de}
\author{Victoria Fernandez Abrevaya}
\orcid{0000-0002-9829-4929}
\affiliation{%
 \institution{Max Planck Institute for Intelligent Systems}
 \city{Tübingen}
 \country{Germany}}
\email{victoria.abrevaya@tuebingen.mpg.de}
\begin{abstract}
Our goal is to efficiently learn personalized animatable 3D head avatars from videos that are geometrically accurate, realistic, relightable, and compatible with current rendering systems. While 3D meshes enable efficient processing and are highly portable, they lack realism in terms of shape and appearance. Neural representations, on the other hand, are realistic but lack compatibility and are slow to train and render. Our key insight is that it is possible to efficiently learn high-fidelity 3D mesh representations via differentiable rendering by exploiting highly-optimized methods from traditional computer graphics and approximating some of the components with neural networks. To that end, we introduce \moniker, a technique that enables the creation of animatable and relightable mesh avatars from a single monocular video. First, we learn a canonical geometry using a mesh representation, enabling efficient differentiable rasterization and straightforward animation via learned blendshapes and linear blend skinning weights. Second, we follow physically-based rendering and factor observed colors into intrinsic albedo, roughness, and a neural representation of the illumination, allowing the learned avatars to be relit in novel scenes. Since our input videos are captured on a single device with a narrow field of view, modeling the surrounding environment light is non-trivial. Based on the split-sum approximation for modeling specular reflections, we address this by approximating the pre-filtered environment map with a multi-layer perceptron (MLP) modulated by the surface roughness, eliminating the need to explicitly model the light. We demonstrate that our mesh-based avatar formulation, combined with learned deformation, material, and lighting MLPs, produces avatars with high-quality geometry and appearance, while also being efficient to train and render compared to existing approaches.
\end{abstract}
\begin{teaserfigure}
\centering
\includegraphics[trim=0 0 0 0, clip=true, width=.95\linewidth]{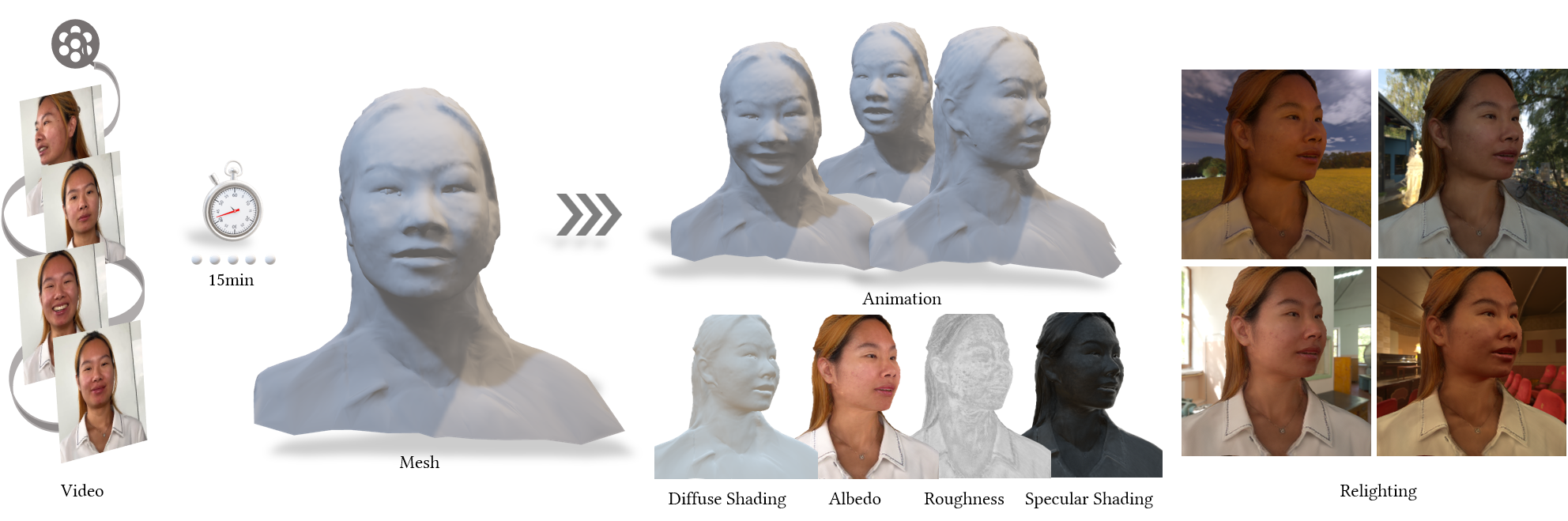}
\vspace{-0.1in}
\caption{We present \moniker{}, a method for rapidly building relightable head avatars from monocular videos. Our method estimates a high-fidelity mesh geometry that can be efficiently animated using learned blendshape and linear-blend-skinning-weight fields. Moreover, we model the intrinsic albedo, roughness, specular reflections, and an indirect representation of the light, enabling relighting in novel scenes.}
\label{fig:teaser}
\end{teaserfigure}
\maketitle
\section{Introduction}
\begin{figure*}[t]
\centerline{\includegraphics[trim=-3em 0em -3em 0em, clip=true, width=\linewidth]{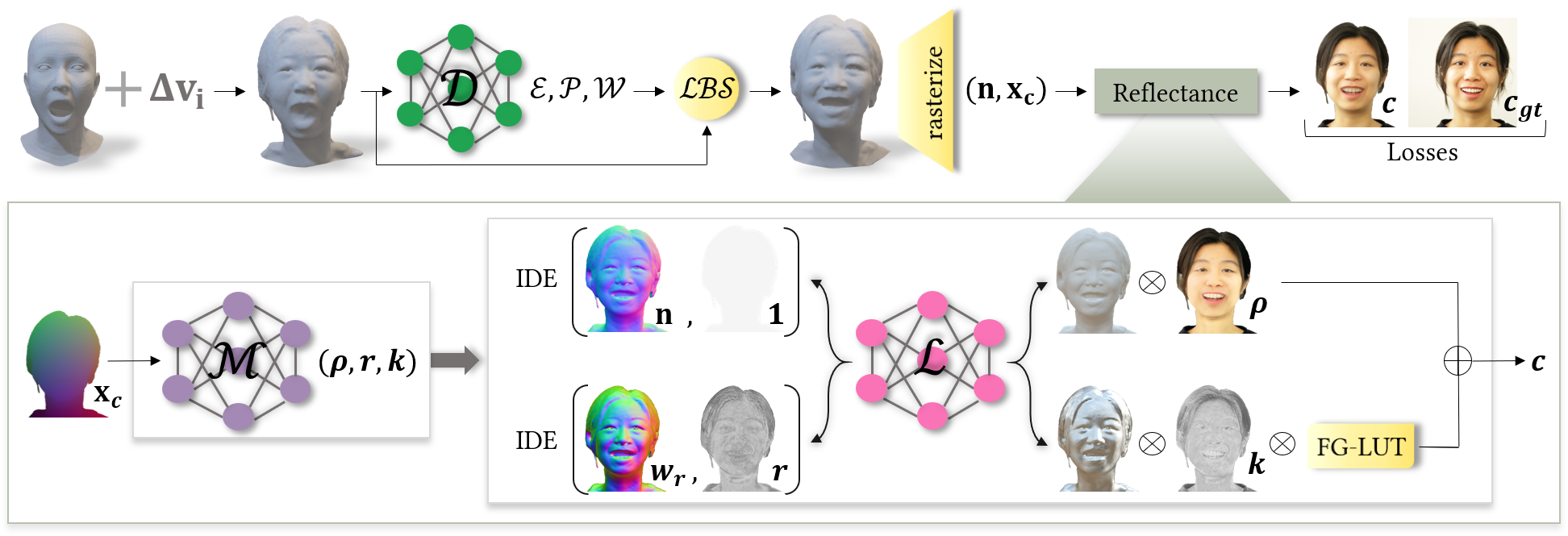}}
\vspace{-0.15in}
\caption{\textbf{Method pipeline.} Top: Given an input video, we optimize for vertex displacements to obtain a canonical geometry. A \emph{deformation} network $\dmlp$ (green) then predicts FLAME~\cite{Li_2017_flame} expression blendshapes $\mathcal{E}$, pose correctives $\mathcal{P}$ and blend skinning weights $\mathcal{W}$ given canonical vertices, which are used to deform the mesh into the corresponding expression and pose. The deformed mesh is rasterized following a deferred shading pipeline to obtain per-pixel canonical coordinates $\xc$ and deformed normals $\n$. Bottom: $\xc$ is used to query the \emph{material} network $\fmlp$ (purple) to obtain the albedo $\albedo$, roughness $\roughness$, and specular intensity $k$. Next, the \emph{lighting} network $\intmlp$ (pink) obtains an estimate of the diffuse shading and specular reflection from the normal and reflection vectors, while taking roughness into account. We use physically-based rendering to compute the final color, which is compared with the ground-truth frame during training. %
}
\label{fig:main_method}
\end{figure*}
\begin{table}[tbh]
\caption{Compared to other methods, \moniker{} converges rapidly, reconstructs high-fidelity geometry, is compatible with graphics pipelines since it employs a mesh representation, and produces head avatars that can be relighted.}
\resizebox{\columnwidth}{!}{%
\begin{tabular}{lcccc}
\hline
Method &
  \begin{tabular}[c]{@{}c@{}}Converges\\ within 15 mins\end{tabular} &
  \begin{tabular}[c]{@{}c@{}}High-fidelity\\ geometry\end{tabular} &
  \begin{tabular}[c]{@{}c@{}}Compatible with\\ graphics pipelines\end{tabular} &
  Relightable \\ \hline
IMavatar    &    \xmark          & \cmark &  \xmark            &    \xmark          \\
NHA         &       \xmark       &        \xmark      & \cmark &      \xmark        \\
PointAvatar &       \xmark       & \cmark &       \xmark       &    \cmark          \\
INSTA       & \cmark &       \xmark       &      \xmark        &    \xmark          \\
FLARE       & \cmark & \cmark & \cmark & \cmark \\ \hline
\end{tabular}%
}
\label{tab:intro-table}
\end{table}
There has been remarkable progress on learning personalized 3D 
facial assets, moving from complex and expensive high-end systems \cite{beeler2011high, debevec2000acquiring, ghosh2011multiview} to using single commodity sensors as input \cite{Zheng_2022_imavatar, Grassal_2022_nha, zielonka2022insta}. Although a quality gap still exists, it is being gradually bridged by neural methods that leverage implicit or explicit shape representations. In particular, signed distance fields \cite{Zheng_2022_imavatar} and point clouds \cite{zheng2023pointavatar} have been used to obtain impressive 3D reconstructions, while NeRF-based \cite{Mildenhall_2020_nerf} approaches \cite{zielonka2022insta, Gafni_2021_nerface} have shown an outstanding ability to synthesize novel views of the subject. Further, these methods are trained such that the learned avatars can be controlled with novel poses and expressions, making them appropriate for entertainment and telecommunication.

There are several challenges that remain for existing head avatars to be widely applicable in industry. First, the majority of methods are slow to train and/or to render, taking hours \cite{zheng2023pointavatar} or days \cite{Grassal_2022_nha, Zheng_2022_imavatar} of processing to obtain a single, scene-dependent avatar. This limits the scope of applications and hinders the creation of immersive experiences. Fast approaches have recently been proposed \cite{Xu2022manvatar, zielonka2022insta, Gao2022nerfblendshape}, but they suffer from low-quality geometry and often do not generalize well to novel views. Second, to achieve high-quality reconstructions, current methods use shape representations that are not compatible with standard graphics pipelines. Ideally, a mesh representation should enable easy asset extraction and integration. 
However, recent neural methods that are built on triangulated meshes \cite{Grassal_2022_nha, Khakhulin_2022_rome} do not achieve the same geometric quality as methods based on more flexible representations. Finally, most neural approaches generate avatars that can only be rendered in the same environment in which they were captured since they do not disentangle the light from intrinsic material properties. What is still missing is an efficient method to extract head avatars that have high-fidelity geometry and can be animated and relit.

In this work we present a new method, \moniker{} (Fast Learning of Animatable and RElightable mesh avatars), for building 3D facial avatars from monocular videos that addresses all of these challenges, as shown in Table~\ref{tab:intro-table} and Figure~\ref{fig:teaser}. We use a mesh representation to allow easy integration as well as fast computation during training and at inference time. We represent the canonical head geometry as a triangular mesh with optimizable vertex locations and learn blendshapes as well as skinning-weight fields to deform the canonical mesh given FLAME~\cite{Li_2017_flame} expression and pose parameters. 
To disentangle the intrinsic material properties and extrinsic light conditions, we leverage physically-based rendering~\cite{cook1982reflectance, walter2007microfacet} where materials and lighting are represented by multi-layer perceptrons (MLPs). Specifically, we use the Disney material model \cite{Burley2012PhysicallyBasedSA} and represent albedo, roughness, and specular intensity as hash-encoded spatial MLPs~\cite{mueller2022instant}. To render color efficiently we adopt the split-sum approximation proposed in \cite{karis2013real}. However, explicitly computing the environment light is challenging with monocular head videos given their narrow field of view.
To address this, we approximate the pre-filtered environment map in the split-sum approximation with a neural network, together with an Integrated Directional Encoding (IDE)~\cite{verbin2022refnerf} that accounts for different roughness levels. Our networks are trained using a two-stage approach, where the first stage is focused on geometry and then the second stage refines the color by leveraging the hash-grid encoding \cite{mueller2022instant}. This allows \moniker{} to control the pace at which geometry and color are learned relative to each other, achieving detailed results in both areas. 
While maintaining high accuracy, our method is carefully designed to improve training and rendering efficiency:
\begin{inparaenum}[(1)]
\item The canonical material estimation MLPs are fueled by hash-grid encoding~\cite{mueller2022instant}, which effectively represents high-resolution mappings with shallow MLPs, boosting query speed significantly;
\item The neural split-sum approximation reduces the evaluation of 
expensive integrals into one look-up in the pre-integrated texture~\cite{karis2013real, munkberg2021nvdiffrec}, as well as one forward pass of an MLP;
\item Our morphable mesh representation enables efficient differentiable rasterization with existing tools~\cite{Laine2020diffrast}, in contrast to implicit representations that require hundreds of queries per pixel.
\end{inparaenum}
Thanks to the above components, our method reconstructs detailed relightable avatars in around 15 minutes. Our experiments show that our proposed approach achieves high-fidelity geometry as well as realistic renderings, which are on par with, or superior to, existing approaches while being much faster to train as demonstrated in Figure~\ref{fig:speed}. Code is available for research purposes at \url{https://flare.is.tue.mpg.de}. 
\section{Related Work}
\subsection{3D head avatars from videos}
 Creating animatable 3D head avatars from videos is a popular research topic because it replaces the need for complex capture equipment \cite{debevec2000acquiring, beeler2011high, riviere2020single, ghosh2011multiview} with more easily accessible commodity sensors. 
 Classic approaches \cite{garrido2016reconstruction, thies2016face2face} employ statistical models \cite{blanz1999morphable} to recover the 3D shape and appearance, but only focus on the facial area and produce relatively coarse reconstructions. 
NerFACE~\cite{Gafni_2021_nerface} was the first to use dynamic neural radiance fields (NeRF) \cite{Mildenhall_2020_nerf} to represent head avatars. 
IMavatar~\cite{Zheng_2022_imavatar} recovers accurate geometry using implicit surfaces by jointly learning canonical head geometry and expression deformations. However, methods based on implicit representations can be inefficient to train and render. PointAvatar~\cite{zheng2023pointavatar} uses a similar deformation model but employs a point cloud representation, enabling faster rasterization and better image quality. Recently, several methods~\cite{Gao2022nerfblendshape, zielonka2022insta, Xu2022manvatar} employ InstantNGP~\cite{mueller2022instant} to speed up radiance field queries and can reconstruct avatars within 5 to 20 minutes. To the best of our knowledge, none of these fast avatar reconstruction methods produce high-quality surface normals. Neural Head Avatar (NHA) \cite{Grassal_2022_nha} reconstructs mesh-based avatars with complete head and hair geometry. However, the reconstructed geometry is relatively coarse, with many  details  represented in the texture space. None of these recent neural methods factorize light and albedo, with the exception of PointAvatar, which performs a rudimentary factorization using a diffuse shading model. In contrast, our method reconstructs mesh-based avatars with high-quality geometry within 15 minutes, and factorizes lighting into albedo, roughness and extrinsic illumination. This enables our avatars to be readily rendered in new scenes.

\subsection{Relightable 3D reconstruction from multi-view images}
The ability to learn relightable 3D assets from 2D observations has extensive applications in AR and VR content creation. Several  previous methods~\cite{bi2020neural, zhang2021physg, boss2021nerd, Zhang2021nerfactor, srinivasan2021nerv, verbin2022refnerf} leverage neural implicit representations
such as NeRF~\cite{Mildenhall_2020_nerf} or neural SDF~\cite{park2019deepsdf, mescheder2019occupancy}, which benefit from unconstrained topology but are inefficient to render. 
Recently, \cite{munkberg2021nvdiffrec} convert neural SDFs to meshes with differentiable marching tetrahedrons~\cite{shen2021dmtet} and employ physically-based rendering to reconstruct high-quality relightable 3D assets in less than an hour. 
Neural-PIL \cite{boss2021neural} leverages a similar idea to us and approximates parts of the split-sum formula \cite{karis2013real} with neural networks. However, the method requires pre-training on a large dataset, which hinders generalization, and is only tested with multi-view images that have full coverage of the scene. 

To obtain 3D head avatars that are both animatable and relightable, recent methods~\cite{feng2022TRUST, dib2021towards} leverage the deformable geometry of pretrained 3DMMs~\cite{paysan2009bfm, Li_2017_flame}, and predict albedo and lighting from a single image. 
SIRA~\cite{caselles2022sira} improves the coarse 3DMM geometry by learning a deformable SDF but requires a large number of 3D scans for training. 
In contrast, our method reconstructs relightable 3D head avatars from a single monocular video, achieving high-quality geometry without requiring expensive 3D scans for prior training.
\section{Method}
\begin{figure}
\centering
\begin{subfigure}{0.23\textwidth}
    \includegraphics[trim=0em 0em 0em 1em, clip=false, width=\linewidth]{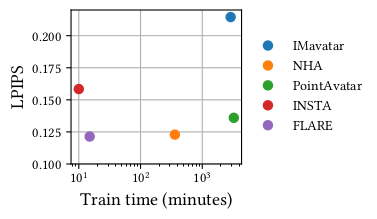}
    \caption{Image quality}
    \label{fig:speed_image}
\end{subfigure}
\begin{subfigure}{0.23\textwidth}
    \includegraphics[trim=0em 0em 0em 1em, clip=false, width=\linewidth]{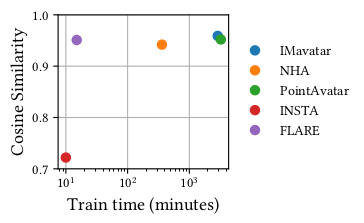}
    \caption{Geometric quality}
    \label{fig:speed_geom}
\end{subfigure}
\vspace{-0.1in}
    \caption{Training time vs image quality (a) and geometric quality (b)  for SOTA methods. Our method is nearly as quick as INSTA while performing on-par or better than competitors. Lower is better for (a) and higher is better for (b). }
    \label{fig:speed}
\end{figure}
Given a fixed-viewpoint video of a person with frames $\{ I_1, \dots, I_N \}$, foreground masks $\{ M_1, \dots, M_N \}$, and pre-computed FLAME \cite{Li_2017_flame} parameters for shape $\flameshape$, expressions $\{ \flameexpr_1, \dots, \flameexpr_N \}$, and poses $\{ \flamepose_1, \dots, \flamepose_N \}$, we jointly train 
\begin{inparaenum}[(1)]
\item the deforming head geometry, parameterized by a canonical mesh with optimizable vertex locations and expression deformation fields (Sec.~\ref{sec:method_geometry});  
\item the intrinsic surface reflectance properties, including albedo, roughness, and specular intensity, represented by an MLP in canonical space (Sec.~\ref{sec:method_reflectance}), 
and \item a \lightmlp{} that approximates the pre-filtered environment map of the scene. 
\end{inparaenum}
To train these we follow a physically-based rendering approach and rasterize the mesh into images, which are compared with the ground-truth frames (Sec.~\ref{sec:training}). Figure~\ref{fig:main_method} gives an overview of the method.
\subsection{Geometry}
\label{sec:method_geometry}
To achieve high train- and test-time efficiency, and for compatibility with standard graphics pipelines, we use triangle meshes as the geometric representation. As shown in Figure ~\ref{fig:main_method}, we learn a single canonical mesh that best explains all views, along with deformation fields that transform the canonical mesh given FLAME pose and expression parameters. We describe each of these below.

\subsubsection{Canonical mesh.} Given a pre-defined FLAME template mesh $\mathcal{V} = (V, \mathcal{F})$, with the set of vertices $V = \{ \vtx_1, \dots, \vtx_M |  \vtx_i \in \real^3 \}$, and the set of triangular faces $\mathcal{F}$, we obtain a personalized shape by optimizing $\{ \Delta \vtx_i |  i=1 \dots M, \Delta \vtx_i \in \real^3 \}$, such that the final canonical mesh vertices are $\{ \vtx_i + \Delta \vtx_i |  i=1\dots M \}$. To facilitate learning we employ a coarse-to-fine approach~\cite{worchel2022_nds, zheng2023pointavatar}, where we upsample the mesh to $\backsim11k$ vertices vertices during training using the algorithm in~\cite{Botsch2004ARA}.

\subsubsection{Deformation field.} We deform the canonical geometry using the FLAME parameters computed during the pre-processing step. Specifically, given a canonical vertex $\vtx \in \real^3$, we deform it as follows:
\begin{eqnarray}
    \lefteqn{FLAME(\vtx, \mathcal{P}, \mathcal{E}, \mathcal{W}, \flamepose, \flameexpr) =} \nonumber \\
    & & LBS( \vtx + B_P(\flamepose; \mathcal{P}) + B_E(\flameexpr; \mathcal{E}), J(\flameshape), \flamepose, \mathcal{W}),
    \label{eq:deformation}
\end{eqnarray}
where $J(\flameshape)$ is the joint regressor, LBS is the standard linear blend-skinning function with blend-skinning weights $\mathcal{W}$, $\flamepose$ and $\flameexpr$ are the FLAME pose and expression parameters, and $B_P(\cdot)$ and $B_E(\cdot)$ compute the pose and expression offsets using the blendshape components $\mathcal{P}$ and $\mathcal{E}$. Similar to IMavatar~\cite{Zheng_2022_imavatar}, we train a deformation network $\dmlp$ parameterized by an MLP that, given a canonical vertex location $\vtx$, returns the expression blendshapes $\mathcal{E} \in \mathbb{R}^{n_{e} \times 3}$,  the pose correctives $\mathcal{P} \in \mathbb{R}^{n_{j} \times 9 \times 3}$, and the linear blend skinning weights $\mathcal{W} \in \mathbb{R}^{n_{j}}$ of the vertex (with $n_e$ and $n_j$ the number of expression parameters and bone transformations, respectively), 
\begin{equation}
    \dmlp(\vtx): \mathbb{R}^3 \rightarrow \mathcal{E, P, W}.
\end{equation}
Note that, while IMavatar requires a costly root-finding process to search for canonical correspondences given deformed ray samples, our mesh formulation avoids this by directly deforming the canonical mesh and rasterizing it. 
\subsection{Reflectance}
\label{sec:method_reflectance}
To make our avatars relightable, we factorize the observed colors into learned albedo, roughness, and specular intensity, as well as a neural representation of the environment illumination. We adopt the Disney shading model~\cite{Burley2012PhysicallyBasedSA} and modify it to better suit our input data. We elaborate on the appearance model in the following. 

\subsubsection{Physically-based rendering.}
According to the classic rendering equation~\cite{kajiya1986rendering}, the radiance $L_o(\x, \wo) \in \real^3$ leaving from a surface point $\x \in \real^3$ with normal vector $\mathbf{n}$ in the direction $\wo$ is modeled as
\begin{equation}
\label{eq:rendering}
    L_o(\x, \wo) = \int_{\Omega} f_r(\x, \wi, \wo) L_i (\wi) (\mathbf{n} \cdot \wi) d\wi,
\end{equation}
where the integral is over the hemisphere $\Omega = \{ \wi | (\wi \cdot \n) > 0  \}$, $f_r(\x, \wi, \wo)$ is the bi-directional reflectance distribution function (BRDF),
and $L_i$ is the incoming light intensity from direction $\wi$.

Following the dichromatic reflection model~\cite{shafer1985_dichromatic}, the BRDF is decomposed into a diffuse term $f_d$ and a specular term $f_s$, and the total reflectance is calculated as $f_r(\x, \wi, \wo) = f_d(\x, \wi, \wo) + k(\x) f_s(\x, \wi, \wo)$, where $k(\x)$ is a spatially-varying specular intensity factor that weighs the contribution of the specular BRDF, similar to \cite{riviere2020single}. The rendering equation then  becomes: 
\begin{equation}
\begin{split}
\label{eq:rendering_diffspec}
    L_o(\x, \wo) = 
        \underbrace{\int_{\Omega} f_d(\x, \wi, \wo) L_i (\wi) (\n \cdot \wi) d\wi}_{L_o^{diff}} + \\
        k(\x) \underbrace{\int_{\Omega} f_s(\x, \wi, \wo) L_i (\wi) (\n \cdot \wi) d\wi .}_{L_o^{spec}}
\end{split}
\end{equation}
We use a simple Lambertian model for the diffuse term: 
\begin{equation}
\label{eq:diffuse}
     L_o^{diff}  = 
     \frac{\albedo(\x)}{\pi} \int_{\Omega} L_i (\wi) (\n \cdot \wi) d\wi,
\end{equation}
where $\rho$ is the spatially-varying RGB albedo. 
For the specular term we use the Cook-Torrance microfacet model \cite{cook1982reflectance}:
\begin{align}
\label{eq:cook_torrance}
    & L_o^{spec} = \\ \nonumber & 
    \int_{\Omega} \frac{D(\n, \wo, \wi, \roughness) G(\wo, \wi, \roughness) F(\wo, \wi, F_0)}{4 (\wo \cdot \n)(\wi \cdot \n)} L_i (\wi) (\n \cdot \wi) d\wi.
\end{align}
Here, the surface roughness $\roughness$ modulates the microfacet normal distribution function $D$, and the geometry attenuation function $G$ that accounts for self-shadowing. 
$F$ denotes the Fresnel equation that describes the proportion of light reflected at different surface angles.  
We follow the Disney material model for the specific choice of $D$, $G$ and $F$ functions, see~\cite{Burley2012PhysicallyBasedSA, karis2013real}. 

\subsubsection{Estimating intrinsic materials. }
We optimize the albedo and roughness of our head model, as well as specular intensity values. 
These properties are canonical properties of the surface and remain constant during facial deformation. Therefore, we employ an MLP, $\fmlp$, that receives canonical surface points $\xc$ as input and predicts albedo $\albedo$, roughness $\roughness$ and specular intensity $k$: 
\begin{equation}
\label{eq:intrinsicmlp}
    \fmlp(\xc) : \mathbb{R}^3 \rightarrow \albedo, \roughness, k.
\end{equation}

\subsubsection{Split-sum approximation.} 
The split-sum approximation~\cite{karis2013real} was proposed to efficiently evaluate the specular reflectance by splitting it into two integrals that can be pre-computed:
\begin{align}
\label{eq:split-sum}
    & L_o^{spec} \approx \\ \nonumber & 
    \int_{\Omega} L_i(\wi) D(\n, \wi, \wo, \roughness) (\wi \cdot \n)  d\wi \int_{\Omega} f(\wi, \wo) (\wi \cdot \n) d\wi.
\end{align}
The first term corresponds to a \emph{pre-filtered environment map}, where the environment light $L_i$ is convolved with the normal distribution function $D$. This term is pre-computed for a set of roughness values and stored as a series of 2D look-up textures (LUT), where each mipmap level is selected based on roughness, and each texture is indexed by the reflection vector $\refl = 2 (\wo \n) \n - \wo$.
The second integral, known as the \emph{BRDF integration map} contains the rest of the terms, and it is equivalent to integrating Equation~\ref{eq:cook_torrance} with a white environment map ($L_i(\wi) = 1$) \cite{karis2013real}. 
This term depends on the roughness $\roughness$ and the cosine angle $(\wo \cdot \n)$, and it is also stored as an LUT, which will be referred to as $FG-LUT$.

\subsubsection{Neural split-sum approximation.}
The split-sum approximation can help to efficiently learn a rich model of illumination, and has been used to disentangle the light and materials from multi-view images \cite{munkberg2021nvdiffrec}. 
However, our setting considers as input a fixed viewpoint video, which is a more challenging scenario for light disentanglement. We found through experiments that optimizing environment maps directly often leads to sub-optimal results (See Figure~\ref{fig:pbr}).
To address this, we approximate the pre-filtered environment map in Eq. ~\ref{eq:split-sum} with a neural network: 
\begin{equation}
    \intmlp(\refl, \roughness) \approx \int_{\Omega} L_i(\wi) D(\n, \wi, \wo, \roughness) (\wi \cdot \n)  d\wi. 
\end{equation}
To design this neural network, we observe that the roughness parameter $\roughness$ influences the output via the normal distribution function $D$, i.e., a larger roughness corresponds to a wider distribution and leads to blurrier filtered light maps.
In 2D LUTs, the pre-filtered environment maps for different roughness values are represented as mipmaps. A key challenge for the neural split-sum approximation is to model this behavior for different roughness levels. 
To address this, we propose to adapt the Integrated Directional Encoding (IDE) \cite{verbin2022refnerf} to represent different mip levels of neural fields. 
The IDE encodes the input reflection vector $\refl$ through the expected value of a set of spherical harmonics under a von Mises-Fischer (vMF) distribution centered at $\refl$, where the concentration parameter $\kappa$ is defined as the inverse roughness $1 / \roughness$:
\begin{equation}
 \label{eq:ide}
    IDE(\refl, \roughness) = 
    \mathbb{E}_{\omega \sim vMF(\refl, 1/r)} [ Y_l^m | (l,m) \in \mathcal{M}_L],
\end{equation}
with $\mathcal{M}_L = \{ (l,m) : l=1 \dots 2^L, m=0 \dots l \}$, $Y_l^m$ the spherical harmonics basis functions, and $L=4$. 
In practice, this positional encoding limits the representational power of the neural network when using larger roughness values, which essentially mimics the behavior of mipmap levels in a continuous manner. Note that the incident illumination $L_i$, now represented as part of the pre-filtered light MLP $\intmlp$, also determines the diffuse shading.  
We observe that setting the roughness to its maximum value $\roughness=1$ within the GGX distribution for $D$ (employed by the Disney material model) equates to $1/\pi$, and the pre-filtered environment map term becomes the diffuse shading of Equation~\ref{eq:diffuse}. 
Hence, we can use $\intmlp$ to represent both the diffuse shading and the specular pre-filtered environment map: 
\begin{align}
    L_o^{diff} & = \frac{\albedo(\x)}{\pi} \cdot \intmlp(IDE(\n, 1))\\
    L_o^{spec} & = \intmlp(IDE(\refl, \roughness)) \cdot FG-LUT(\roughness, \wo \cdot \n).
\end{align}
We thus replace the explicit integration of a scene environment map with a single query over the pre-filtered light MLP, 
while still grounding the formulation on a physics-based model via the $FG-LUT$ term. At test time we relight the avatar by simply replacing $\intmlp$ with a pre-filtered environment map.

\subsubsection{Color prediction.} The final outgoing radiance is calculated as
\begin{align}
\label{eq:final-color}
     L_o(\x, \wo) &= \frac{\albedo(\x)}{\pi} \cdot \intmlp(IDE(\n, 1)) + \nonumber \\
        &k(\x) \intmlp(IDE(\refl, \roughness)) \cdot FG-LUT(\roughness(\x), \wo \cdot \n).
\end{align}
\section{Training}
\label{sec:training}
\subsection{Loss Functions} 
In this section, we discuss the loss functions employed by \moniker, grouped by image-related losses, geometry-related losses, de\-form\-ation-related losses, and material regularizers. 

\subsubsection{Image}
Given a ground-truth frame $I_j$ and a predicted image $\tilde{I}_j$, we compute (1) the $L2$ loss in log space between the masked ground-truth and the predicted image following \cite{munkberg2021nvdiffrec}:
\begin{equation}
 \mathcal{L}_{RGB} = || log(I_j) - log(\tilde{I}_j) ||_2^2, 
\end{equation}
(2) an L2 loss between ground-truth and predicted binary masks: 
\begin{equation}
    \mathcal{L}_{\mathit{mask}} = || M_j - \tilde{M}_j ||_2^2,
\end{equation}
and (3) a perceptual loss \cite{Johnson2016Perceptual} given as: 
\begin{equation}
\mathcal{L}_{vgg} = || F_{vgg}(I_j) - F_{vgg}(\tilde{I}_j) ||_2^2,
\end{equation}
where $F_{vgg}$ represents the extracted features from the first four layers of a pre-trained VGG~\cite{Simonyan_2014_vgg} network.

\subsubsection{Geometry} 
During the optimization of mesh vertices, it is necessary to constrain them in order to avoid self-intersections and to obtain a coherent shape. We follow \cite{worchel2022_nds, luan2021unified} and use a Laplacian smoothness regularizer where the magnitude of the differential coordinates of each vertex is minimized. For canonical vertices given by $\{ \vtx_i + \Delta \vtx_i |  i=1\dots M \}$, the regularizer is defined as:
\begin{align}
    \mathcal{L}_{\mathit{laplacian}} &= \frac{1}{M} \sum_1^M ||\delta_i||_2^2
\end{align}
 where $\delta_i = (LV)_i$ are the differential coordinates of the $i$-th vertex, and $L \in \real^{M \times M}$ the graph Laplacian of the mesh \cite{sorkine2005laplacian}. Additionally, we apply a normal consistency term \cite{worchel2022_nds, luan2021unified} that enforces cosine similarity between neighboring face normals and is given by:
\begin{equation}
    \mathcal{L}_{normal} = \frac{1}{|\mathcal{F}|} \sum_{(i,j) \in \mathcal{F}} (1-\n_i \cdot \n_j)^2. 
\end{equation}
$\mathcal{F}$ is the set of triangle pairs that share an edge, and $n_i$ is the normal of triangle $i$. 

\subsubsection{Deformation}
We regularize the blendshapes and skinning weights similar to \cite{Zheng_2022_imavatar} as follows:
\begin{eqnarray}
       \lefteqn{\mathcal{L}_{\mathit{flame}} =} \\
       &&\frac{1}{M} \sum_1^M (  \lambda_e ||\mathcal{E}_i - \hat{\mathcal{E}}_i||_2 +
        \lambda_p||\mathcal{P}_i - \hat{\mathcal{P}}_i||_2 + 
        \lambda_w||\mathcal{W}_i - \hat{\mathcal{W}}_i||_2
        ),\nonumber
\end{eqnarray}
where $\mathcal{L}_{flame}$ regularizes $\mathcal{E}$, $\mathcal{P}$ and $\mathcal{W}$ with pseudo ground-truth $\hat{\mathcal{E}}$, $\hat{\mathcal{P}}$ and $\hat{\mathcal{W}}$, obtained from the nearest vertex of the FLAME~\cite{Li_2017_flame} template. Here, $\lambda_e = 50, \lambda_p = 50, \lambda_w = 2.5$. \\

\subsubsection{Material Regularization}
We apply a white light regularization over the diffuse shading as in \cite{munkberg2021nvdiffrec}:
\begin{equation}
    \mathcal{L}_{\mathit{light}} = \frac{1}{3} \sum_{i=0}^{3} | \overline{c}_i - \frac{1}{3} \sum_{i=0}^3 \overline{c}_i |,
\end{equation}
where $\overline{c}_i$ is the per-channel average intensity. Additionally, we regularize the specular intensity $k$ by computing the z-score of our predicted specular intensities relative to a Gaussian distribution based on the MERL / ETH Skin Reflectance Database \cite{weyrich2006analysis}.  The dataset provides specular intensity measurements for 156 faces, with a mean value of $0.3753$ and a standard deviation of $0.1655$. The regularization is defined as:
\begin{equation}
    \mathcal{L}_{\mathit{spec}}(\xc) = \frac{ \xc - 0.3753 }{0.1655}.
\end{equation}
We employ a similar strategy to regularise the roughness. However, since the statistics reported in \cite{weyrich2006analysis} are computed for the Torrance-Sparrow model, we empirically set the mean to $\mu_{rough} = 0.5$ and standard deviation to $\sigma_{rough} = 0.1$ through visual evaluation. We provide an ablation study in Section~\ref{ablation_loss_functions} to support the choice of this hyper-parameter. The loss is defined as:
\begin{equation}
    \mathcal{L}_{r}(\xc) = \frac{ \xc - \mu_{\mathit{rough}} }{\sigma_{\mathit{rough}}}.
\end{equation}
Finally, we enforce a smoothness constraint for both albedo and roughness similar to \cite{munkberg2021nvdiffrec}, with an additional robust term \cite{barron2019robust} that helps preserve high-frequency details. Specifically, for each canonical point $\xc \in \real^3$ we compute a random displacement vector $\epsilon \in \real^3$ sampled from a Gaussian distribution, and compute the albedo ($\albedo$) and roughness ($\roughness$) for both points. We apply an L1 loss between these two to enforce smoothness within neighboring points as follows:
\begin{align}
    \mathcal{L}_{smooth}(\xc) = f_{\mathit{robust}}(|| \albedo(\xc) - \albedo(\xc+\epsilon) ||_1 ) \\ 
    + f_{\mathit{robust}}( || \roughness(\xc) - \roughness(\xc+\epsilon) ||_1 )
\end{align}
where $f_{\mathit{robust}}$ is the adaptive robust loss function of \cite{barron2019robust}.

\subsubsection{Loss function} The full loss function is as follows:
\begin{equation}
\begin{split}
    \mathcal{L} = 
       \lambda_{RGB} \mathcal{L}_{RGB} + \lambda_{vgg} \mathcal{L}_{vgg} + \lambda_{\mathit{mask}} \mathcal{L}_{\mathit{mask}} + \\
        \lambda_{\mathit{flame}}\mathcal{L}_{\mathit{flame}} + 
        \lambda_{\mathit{laplacian}} \mathcal{L}_{\mathit{laplacian}} + \lambda_{\mathit{normal}} \mathcal{L}_{\mathit{normal}} + \\
        \lambda_{\mathit{smooth}} \mathcal{L}_{\mathit{smooth}} + \lambda_{r} \mathcal{L}_{r} + \lambda_{\mathit{spec}} \mathcal{L}_{\mathit{spec}} + \lambda_{\mathit{light}} \mathcal{L}_{\mathit{light}}
\end{split}
\end{equation}
where $\{ \lambda_i  \in \real \} $ weigh the importance of the corresponding terms and we empirically determined them as: $\lambda_{RGB}=1.0, \lambda_{vgg}=0.1, \lambda_{\mathit{mask}}=2.0, \lambda_{\mathit{flame}}=5.0, \lambda_{\mathit{laplacian}}=60.0,
\lambda_{\mathit{normal}}=0.1,
\lambda_{\mathit{smooth}}=0.01, \lambda_{r}=0.01, \lambda_{\mathit{spec}}=0.01, \lambda_{\mathit{light}}=0.01$. 

\subsection{Training Details}
We train \moniker{} using differentiable rendering to compare the predicted images with ground-truth frames. 
Given the current canonical mesh $\{ \vtx_i + \Delta \vtx_i | i=1 \dots M \}$, we first estimate expression blendshapes $\mathcal{E}$, pose correctives $\mathcal{P}$ and blend skinning weights $\mathcal{W}$ through a forward pass of the deformation network, $\dmlp (\vtx_i + \Delta \vtx_i) \rightarrow (\mathcal{E}, \mathcal{P}, \mathcal{W})$. With the expression and pose parameters $\flameexpr, \flamepose$, we obtain deformed vertex positions $\tilde{\vtx}_i$ using the FLAME function in  Equation~\ref{eq:deformation}, $\tilde{\vtx_i} = FLAME(\vtx_i + \Delta \vtx_i, \mathcal{E}, \mathcal{P}, \mathcal{W}, \flamepose, \flameexpr)$. Following a deferred shading pipe\-line, the deformed vertices are rasterized to obtain triangle indices and barycentric coordinates per pixel. We then interpolate and obtain the corresponding canonical point locations $\xc$, deformed point locations $\xd$ (used to compute $\wo$), and deformed normals $\nd$ for each pixel.
Next, we compute material properties by querying the material network $\fmlp(\xc) \rightarrow (\albedo, \roughness, k)$ (Equation~\ref{eq:intrinsicmlp}). Finally, we query the \lightmlp{}  $\intmlp(\refl, \roughness)$ using the deformed normals $\nd$ to obtain the left-hand side of Equation~\ref{eq:split-sum} and the diffuse shading of Equation~\ref{eq:diffuse}. The final color for the pixel is computed using Equation~\ref{eq:final-color}. 
Our framework is implemented in PyTorch and trained using a single A100 Nvidia GPU with 80GB of memory and a batch size of 4 images per iteration. 

\subsubsection{Two-stage training.}
To learn high-frequency facial features and to enable fast rendering, we incorporate hash-grid encoding~\cite{mueller2022instant} for the material MLP, $\fmlp$. 
However, we found that this approach overfits to colors quickly, learning texture much faster than the geometry, resulting in smoother shapes of lower quality. 
To address this, we design a two-stage training approach. During the first stage, we equip the material MLP with positional encoding~\cite{Mildenhall_2020_nerf} and jointly optimize the geometry $\Delta \vtx_i$, deformation $\dmlp$, material $\fmlp$, and lighting $\intmlp$. The first stage can achieve detailed geometry but often learns blurry texture. During the second stage of training, we leverage the pre-trained mesh geometry and deformation from the previous stage and re-optimize both material and \lightmlp s, where $\fmlp$ is now equipped with high-resolution hash-grid encoding~\cite{mueller2022instant}. With the proposed two-stage training, our method can achieve both high-fidelity geometry and realistic texture (See Fig. ~\ref{fig:hash_pos_both}).
\begin{figure}
\centering
\newcommand{\mywidth}{0.23}
\includegraphics[page=1]{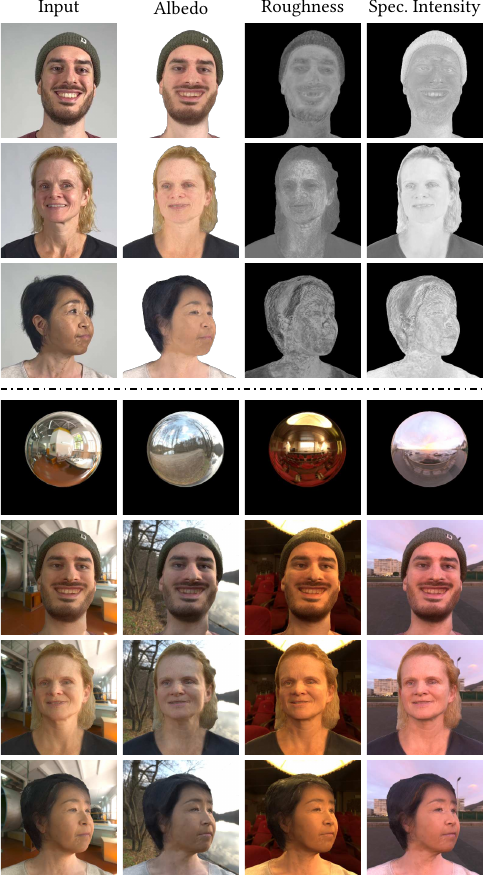}
\caption{Qualitative results. The first three rows illustrate our intrinsic material estimates (albedo, roughness, and specular intensity) for three different subjects. The next three rows show the above subjects in the same pose and expression under 4 different environment maps. 
}
\label{fig:predictions}
\end{figure}
\begin{figure*}
\centering
\newcommand{\mywidth}{0.1}
\includegraphics[page=1]{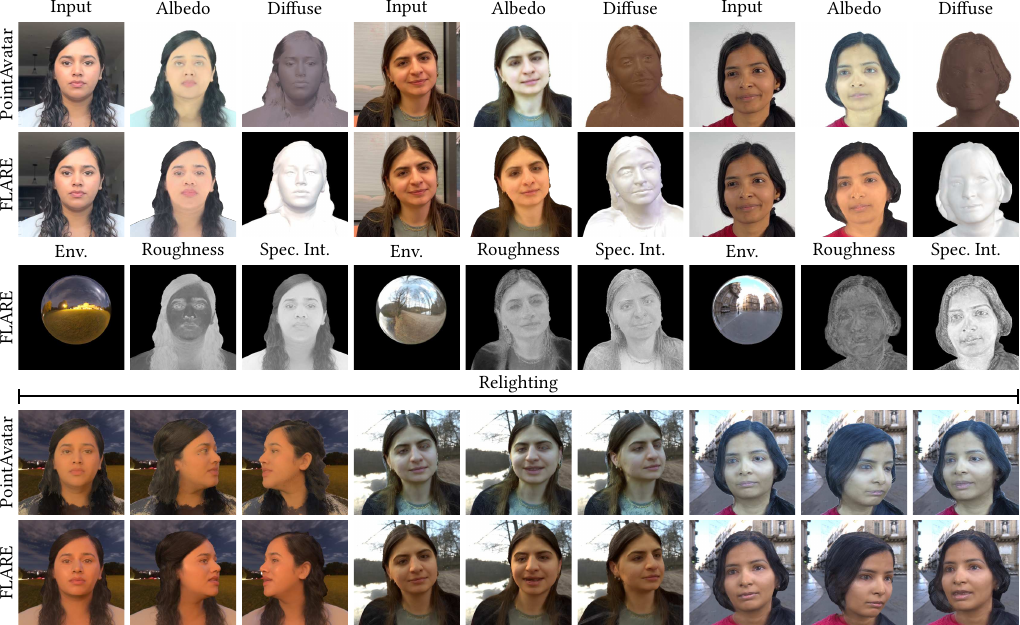}
\caption{Qualitative comparision with PointAvatar. The first two rows show albedo and diffuse shading estimated by \moniker{}  compared with PointAvatar \cite{zheng2023pointavatar}. The next row shows the roughness and specular intensity (Spec.~Intensity) estimated by \moniker{} for the same subjects as above. The bottom two rows contain relighting results of \moniker{} and PointAvatar for the same subjects, animated with test poses and expressions. 
}
\label{fig:cmp_pointavatar}
\end{figure*}
\begin{figure*}
\centering
\newcommand{\mywidth}{0.112}
\includegraphics[page=1]{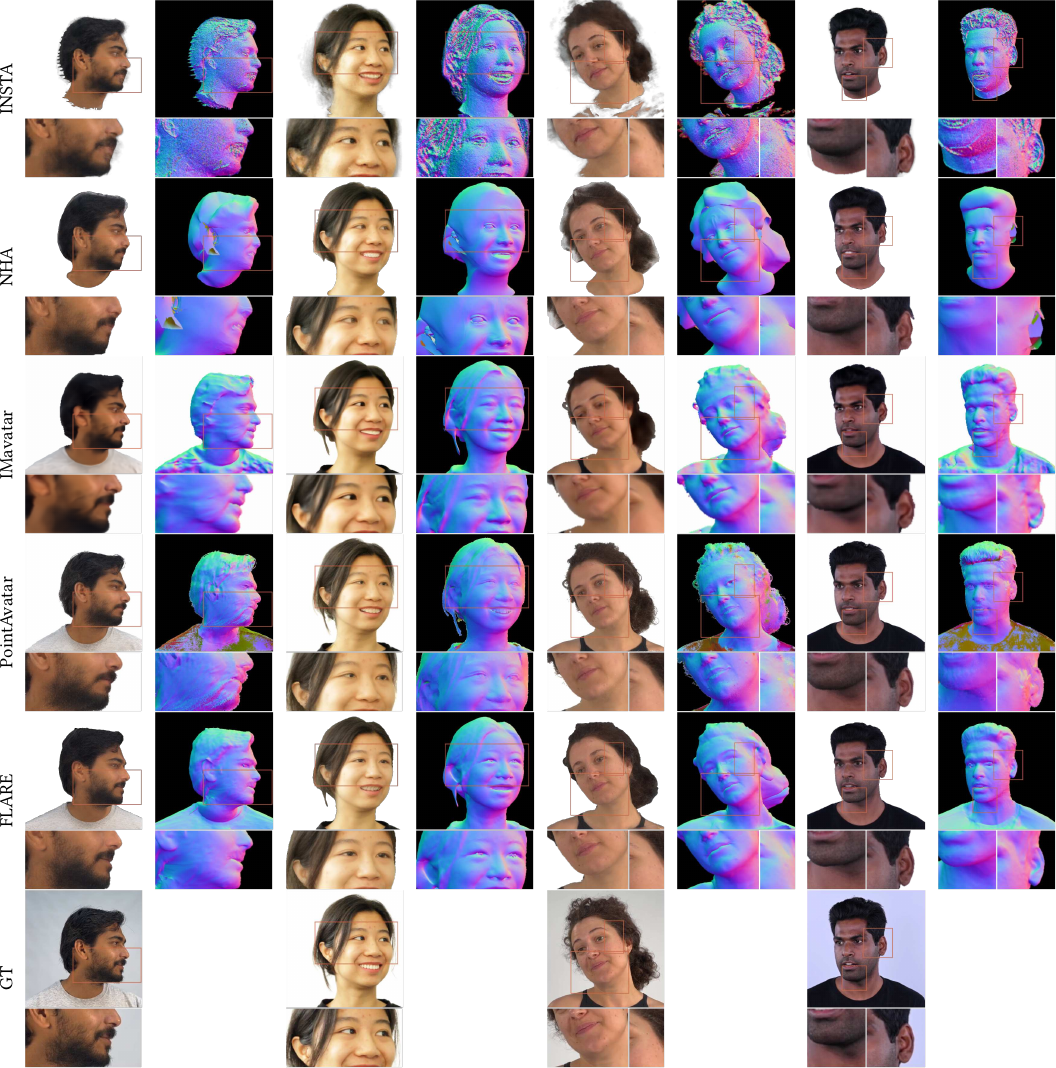}
\caption{Qualitative comparison between \moniker{} and state-of-the-art methods. The canonical representation of each baseline method is animated using test poses and expressions.  Odd columns: generated images. Even columns: generated normals. 
} 
\label{fig:real_qualitative}
\end{figure*}
\begin{figure*}
\centering
\newcommand{\mywidth}{0.12}
\includegraphics[page=1]{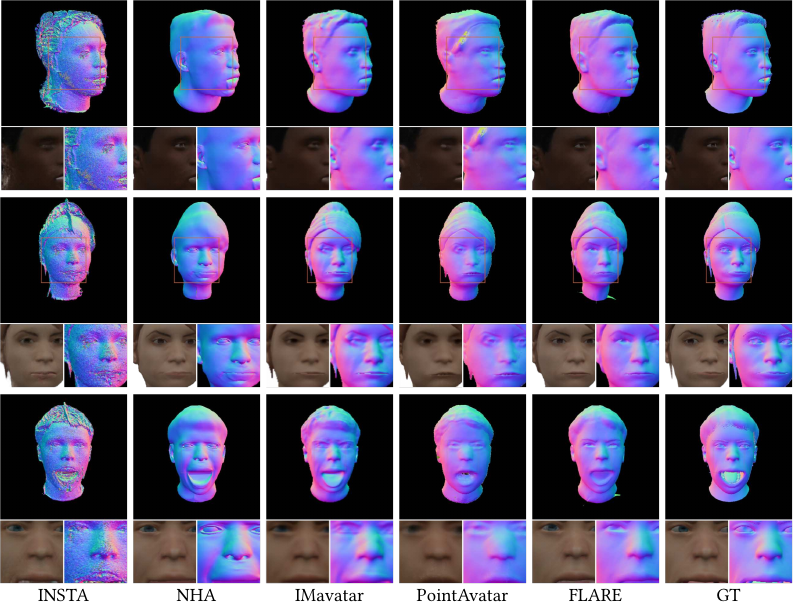}
\caption{Qualitative comparisons on synthetic data. The estimated surface normals and texture on synthetic data are compared with state-of-the-art methods by animating the canonical representation using test poses and expressions. Our method can capture high-fidelity geometry as well as color. GT = Ground Truth. 
}
\label{fig:makehuman}
\end{figure*}
\section{Evaluation}
In this section, we present qualitative and quantitative results of \moniker. First, we show qualitative examples of the individual components, including geometry, albedo, roughness, diffuse and specular shading, as well as relit images (Sec.~\ref{sec:qual_results}). Next, we compare our results with the state-of-the-art (SOTA) baselines in terms of image quality and albedo, as well as geometric accuracy (Sec.~\ref{sec:sota}). Finally, we conduct an ablation study to evaluate our design choices in Sec.~\ref{sec:ablation}. All the results in this section are generated using frames from the test set. For each test frame, we obtain FLAME parameters (pose and expression) from the pre-processing step and animate the personalized canonical representation (geometry) estimated by each baseline method. These animated renderings are relit with novel environment maps. We include additional results in the supplementary video.

\subsection{Dataset}
We use 2 subjects released by \cite{Zheng_2022_imavatar}, 2 by \cite{zheng2023pointavatar} (where 1 subject is captured by a webcam), and 1 by \cite{zielonka2022insta}. We additionally capture 15 subjects with a smartphone to demonstrate the robustness of FLARE for diverse skin tones and shapes. We follow the protocol of \cite{Zheng_2022_imavatar, zheng2023pointavatar} for the capture and obtain an average of 3000-4000 frames for training and around 1000-3000 frames for testing. These new subjects gave prior informed written consent for their data to be used for academic research purposes. In total, we conduct the evaluations for 20 subjects. To measure geometric accuracy we use a  dataset of synthetic heads \cite{Grassal_2022_nha, makehuman}, in which each head has 200 frames for training and 200 frames for testing. 

\subsection{Qualitative Evaluation: Intrinsic Materials and Relighting}
\label{sec:qual_results}
The intrinsic material properties (albedo, roughness, and specular intensity) and relit faces are visualized in Figures \ref{fig:predictions} and \ref{fig:cmp_pointavatar}. The rendered albedo images in Figure~\ref{fig:predictions} illustrate that \moniker{} is capable of removing evident shadows and specular highlights in the face region;  e.g.,~see the subject in the second row. The influence of the predicted roughness values can be visualized in the relit images: the teeth of the subject in the first row and the hair of the subject in the second row are correctly predicted as shiny surfaces (lower roughness values), which results in realistic reflections when relit with new environment lighting. Finally, the robustness of \moniker{} is demonstrated across different skin tones, skin textures, hair types,  hair styles, facial hair, and even accessories such as a cap. Despite having a single monocular video as input, \moniker{} computes geometries and materials that enable realistic and plausible relighting. 

\subsubsection{Comparison with PointAvatar} To the best of our knowledge, PointAvatar \cite{zheng2023pointavatar} is the only other neural avatar method trained from a monocular video that disentangles diffuse shading from albedo. Thus, we qualitatively evaluate the albedo and shading of \moniker{} in Figure~\ref{fig:cmp_pointavatar} by comparing it with PointAvatar. We relight the renderings of PointAvatar using a Lambertian shading model, where we use the predicted surface normals to obtain diffuse shading. From Figure~\ref{fig:cmp_pointavatar}, we observe that the albedo estimated by PointAvatar is biased towards light skin tones and fails to capture the skin color of the subjects. 
In comparison, the albedo estimated by \moniker{} resembles the color of the subject, and much of the shading is removed. 

Relighting \moniker{ }'s estimated materials results in natural looking images. This is due, in part, to the estimated specular highlights, which are absent in PointAvatar's formulation. The specular highlights are visible in the 5th row of Figure~\ref{fig:cmp_pointavatar}, on the left and right cheeks of the first subject from the left, and in the rightmost subject, who has smooth and shiny hair that reflects the environment's light.

\subsection{Comparisons with State-of-the-Art}
\label{sec:sota}
In this section, we compare the results of \moniker{} with the following state-of-the-art (SOTA) methods for neural head avatar estimation from videos: (1) IMavatar \cite{Zheng_2022_imavatar} and (2) PointAvatar \cite{zheng2023pointavatar}, which use  a  deformation module similar to ours, with a signed distance function (SDF) and point cloud representation for geometry, respectively; (3) NHA \cite{Grassal_2022_nha}, which employs a mesh representation along with an alternating training strategy between geometry and color;  and
(4) INSTA \cite{zielonka2022insta}, which learns an animatable avatar using a NeRF~\cite{Mildenhall_2020_nerf} representation and leverages the InstantNGP framework ~\cite{mueller2022instant} for faster optimization. NHA and PointAvatar employ test-time optimization of the expression and pose parameters due to noisy pre-processing estimates. Hence, we report the optimized quantitative evaluations for both NHA and PointAvatar to retain their best performance. However, it must be noted that the reported results of \moniker{}, IMavatar, and INSTA are \emph{not} optimized at test time. \\

\subsubsection{Image quality} First, we compare \moniker{} with SOTA methods with respect to image quality. We use the same FLAME parameters sampled from the test set on all baselines\footnote{INSTA uses a different pre-processing pipeline and the estimated FLAME parameters are different, see Appendix~\ref{ap:preprocessing}. However, the test frames and conveyed expressions remain the same.} and measure the accuracy against the ground-truth frames by using mean absolute error (L1 distance), PSNR, structural similarity index measure (SSIM) \cite{SSIM}, and perceptual loss (LPIPS) \cite{zhang18lpips}. The evaluations are computed only on the masked regions for all the methods.  
Qualitative results can be found in Figure~\ref{fig:real_qualitative}, and quantitative results are shown in Table~\ref{tab:baseline-table}.
\begin{table}[tbh]
\caption{Quantitative comparisons in terms of image quality on real data. The evaluations are performed only on the face region. Red color indicates the best value, yellow second best, and light yellow is the third best.  
}
\begin{tabular}{lcccc}
\hline
 & L1 $\downarrow$ & LPIPS $\downarrow$ & SSIM $\uparrow$ & PSNR $\uparrow$ \\ \hline
IMavatar & 0.0290 & 0.2091 & \cellcolor[HTML]{FFCCC9}0.8491 & 23.0975 \\
NHA & \cellcolor[HTML]{FFFFC7}0.0265 & \cellcolor[HTML]{FFFC9E}0.1243 & 0.8390 & 22.7071 \\
PointAvatar & \cellcolor[HTML]{FFCCC9}0.0234 & \cellcolor[HTML]{FFFFC7}0.1400 & \cellcolor[HTML]{FFFFC7}0.8391 & \cellcolor[HTML]{FFFC9E}24.6520 \\
INSTA & 0.0290& 0.1607 & 0.8379 &  \cellcolor[HTML]{FFFFC7}23.6279 \\ \hline
Ours & \cellcolor[HTML]{FFFC9E}0.0245 & \cellcolor[HTML]{FFCCC9}0.1225 & \cellcolor[HTML]{FFFC9E}0.8421 & \cellcolor[HTML]{FFCCC9}24.7845 \\ \hline
\end{tabular}
\label{tab:baseline-table}
\end{table}
We make the following observations in comparison to prior art: 
(1) In terms of image quality, the baselines perform approximately on par with each other, with \moniker{} obtaining the highest score over half the metrics and second highest over the other half. (2) Despite PointAvatar's ability to capture high-frequency texture details, the point cloud representation, containing approximately 400k points, is susceptible to sparsity at extreme jaw or neck poses. From Figure~\ref{fig:real_qualitative}, 4th row and 5th column, we can observe the artifacts that occur at extreme poses, producing a salt-and-pepper-like noise. 
In comparison, our mesh representation inherently solves the sparsity issue with approximately 11k vertices (we evaluate mesh resolution in Sec.~\ref{sec:meshres}). (3) \moniker{} is able to capture high-frequency texture details better than IMavatar and this is evidenced qualitatively as well as quantitatively, where IMavatar has the weakest LPIPS score. (4) INSTA can converge quickly and produces visually convincing expressions and high-quality texture with forward-facing poses. However, at extreme neck poses we observe noisy texture, possibly due to the volumetric representation that fails to extrapolate well. (5) NHA also employs a mesh-based representation to learn the geometry. However, the predicted mesh is unable to capture high-fidelity details as well as the  baselines and produces an over-smoothed representation. We believe that this is a result of their training strategy in which the geometry is primarily supervised with pseudo-normals from \cite{Abrevaya_2020_CVPR}, unlike the rest of the baselines, which learn geometry exclusively via inverse rendering. Instead,  we carefully consider how fast the texture network is trained compared to the geometry network, and we observe that this was helpful in achieving high-fidelity geometry. We evaluate our training strategy further in Sec.~\ref{sec:ablation}. 

\subsubsection{Geometric Accuracy} 
\label{sec:results_geometry}

\begin{table}[tbh]
\caption{Quantitative comparisons in terms of geometric accuracy on a synthetic dataset. Showing cosine similarity compared to ground-truth normals (higher is better). Red color indicates the highest value, yellow second highest and light yellow is third.}
\begin{tabular}{lrrrr}
\hline
 & \multicolumn{1}{c}{Female 1} & \multicolumn{1}{c}{Female 2} & \multicolumn{1}{c}{Male 1} & \multicolumn{1}{c}{Male 2} \\ \hline
IMavatar & \cellcolor[HTML]{FFCCC9}0.961 & \cellcolor[HTML]{FFCCC9}0.966 & \cellcolor[HTML]{FFCCC9}0.954 & \cellcolor[HTML]{FFFC9E}0.955 \\
NHA & 0.94 & 0.95 & 0.94 & 0.94 \\
PointAvatar & \cellcolor[HTML]{FFFC9E}0.954 & \cellcolor[HTML]{FFFFC7}0.954 & \cellcolor[HTML]{FFFFC7}0.944 & \cellcolor[HTML]{FFCCC9}0.958 \\
INSTA & 0.665 & 0.751 & 0.757 & 0.713 \\ \hline
Ours & \cellcolor[HTML]{FFFFC7}0.950 & \cellcolor[HTML]{FFFC9E}0.955 & \cellcolor[HTML]{FFFC9E}0.948 & \cellcolor[HTML]{FFFFC7}0.953 \\ \hline
\end{tabular}
\label{tab:makehuman}
\end{table}
We quantitatively evaluate the geometry quality on synthetic heads using the renderings generated by the authors of NHA with the open source \textit{MakeHuman} project \cite{makehuman}. Geometric accuracy is measured using the cosine similarity between the ground truth and predicted normals. Results are shown in Table~\ref{tab:makehuman} and Figure~\ref{fig:makehuman}.
IMavatar and \moniker{} exhibit high-fidelity normals that resemble the input identity and obtain relatively close scores quantitatively. However, IMavatar is approximately 200 times slower to train than \moniker{}, mainly due to the root-finding step during ray tracing between deformed and canonical points. Moreover, it uses an SDF representation that requires a post-processing step to obtain a mesh, while \moniker{} can be trained in approximately 15 minutes and directly produces a canonical mesh that can be animated. INSTA, on the other hand, exhibits noisy shapes that can be observed in both Figure~\ref{fig:makehuman} and Figure~\ref{fig:real_qualitative}, and the normals of NHA do not completely capture the identity. 
Note that the synthetic heads have smooth geometry and, consequently, most methods do well with only small numerical differences between methods. 

\subsubsection{Training Time}
Figure~\ref{fig:speed} plots the training time of each method against image quality (LPIPS) and geometric quality (cosine similarity). The plot is measured over the same data as Tables~\ref{tab:baseline-table} and~\ref{tab:makehuman}.  We find that \moniker{} can be trained almost as quickly as INSTA but with better performance in terms of image quality and state-of-the-art results in terms of geometry.

\subsection{Ablation Study}
\label{sec:ablation}
\subsubsection{Loss Functions} 
\label{ablation_loss_functions}  
We evaluate the contribution of the terms in the loss function that are not adopted by prior avatar methods but are crucial in our setting. 

\paragraph{Specular Intensity Regularization} $\mathcal{L}_{spec}$: 
\begin{figure}
\centering
\newcommand{\mywidth}{0.23}
\includegraphics[page=1]{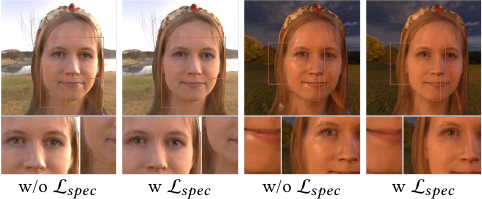}
\caption{Ablation of $\mathcal{L}_{spec}$. Qualitative comparison of relighting results with and without the specular intensity regularizer. Results indicate that it is necessary to constrain the specular intensity statistically to avoid unrealistically sharp highlights.
}
\label{fig:ablations_wo_k}
\end{figure}
The specular intensity $k$ controls the intensity of the specular highlights. In Figure~\ref{fig:ablations_wo_k} we qualitatively evaluate the effectiveness of using the regularizer and show relighting results for one subject with and without the specular intensity regularization. We observe the occurrence of unnaturally sharp highlights that have high intensity around the subject's lower lip and cheek regions when specular intensity is left unconstrained. Constraining it with $\mathcal{L}_{spec}$ makes the non-Lambertian effects more subtle and natural. 

\paragraph{Roughness Regularization $\mathcal{L}_r$:}
\begin{table}[tbh]
\caption{Ablation of $\mathcal{L}_r$. Influence of $\mathcal{L}_{r}$ on image quality when using different mean roughness values.}
\begin{tabular}{ccccc}
\hline
\begin{tabular}[c]{@{}c@{}}Mean\\ Roughness\end{tabular} & L1 $\downarrow$ & LPIPS $\downarrow$ & SSIM $\uparrow$ & PSNR $\uparrow$ \\ \hline
0.3 & 0.0257 & 0.1014 & 0.8772 & 24.341 \\
0.4 & 0.0250 & 0.1031 & 0.8744 & 24.268 \\
0.5 & \bf{0.0239} & \bf{0.0941} & 0.8834 & 24.847 \\
0.6 & 0.0225 & 0.0954 & \bf{0.8824} & \bf{25.023} \\
0.7 & 0.0247 & 0.0985 & 0.8790 & 24.471 \\ \hline
\begin{tabular}[c]{@{}c@{}}w/o\\ regularization\end{tabular} & 0.0265 & 0.1028 & 0.8738 & 23.984 \\ \hline
\end{tabular}
\label{tab:roughness-reg}
\end{table}

\begin{figure}
\centering
\newcommand{\mywidth}{0.23}
\includegraphics[page=1]{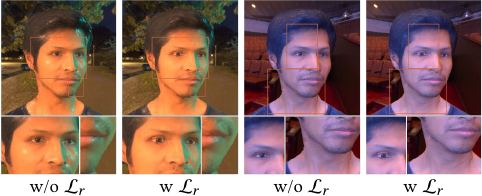}
\caption{Ablation of $\mathcal{L}_r$. Qualitative comparison of relighting results with and without the roughness regularizer. Results indicate that it is necessary to constrain the roughness to ensure non-Lambertian reflections on the skin look  plausible. }
\label{fig:ablations_wo_rr}
\end{figure}
To regularize the roughness we employ a statistical approach similar to specular intensity. However, we know of no suitable database of statistical values for roughness 
that can be used to regularize the appearance model. 
Hence, we employ an empirical mean with a fixed standard deviation of $0.1$, and evaluate the results of using different mean values in Table~\ref{tab:roughness-reg}. Additionally, we evaluate the results of not using this regularization and show qualitative results in Figure~\ref{fig:ablations_wo_rr}. We observe a similar behavior as with specular intensity when roughness is left unconstrained. The final numerical prediction of each subject is not affected by a large margin since the network learns to compensate for wrong predictions with other estimations. 
However, Figure \ref{fig:ablations_wo_rr} reveals that the regularizer helps produce visually realistic renderings. \\

\subsubsection{Standard PBR vs. \Lightmlp}
\begin{figure}
\centering
\newcommand{\mywidth}{0.2}
\includegraphics[page=1]{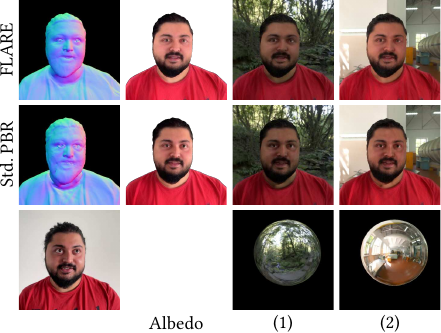}
\caption{Ablation Study, Comparison against learning a full environment map (``standard PBR''), as in \cite{munkberg2021nvdiffrec}. Using this representation typically results in noisier geometry and color. From left to right: predicted geometry, predicted albedo, relighting under two different environment maps. The bottom row shows the input test image, and the two environment maps.}
\label{fig:pbr}
\end{figure}
We compare our proposed approach with a method that estimates a standard texture-based environment map for training, which will be referred to as ``Standard PBR''. For this experiment we use the same hyper-parameters, loss functions, training protocol, and geometric representation of our method, and replace the pre-filtered light MLP $\intmlp$ with a learnable texture of the environment map, where the integral is solved with the approach proposed by \cite{munkberg2021nvdiffrec}. In Figure~\ref{fig:pbr} we visualize an example of geometry and relighting obtained with both methods.  We observe that the standard PBR results in noisy texture and geometry predictions that are evident after relighting the subject. This is probably due to the redundant calculations of the regions in the environment map that are never observed in our monocular setting, creating instability in the optimization process. Further, we can observe that the input image in Figure~\ref{fig:pbr} (bottom left) is captured such that the main light source is from the right of the subject. However, the environment maps have the main light source coming from the left. Here, PBR exhibits shadows in the texture that are retained from the original input data; for instance, see the shadowing on the nose. This is also observed in the estimated albedo, and it is not prominent in our results. 

\subsubsection{Two-stage training}
\begin{figure}
\centering
\newcommand{\myheight}{1.6cm}
\newcommand{\mywidth}{0.3}
\includegraphics[page=1]{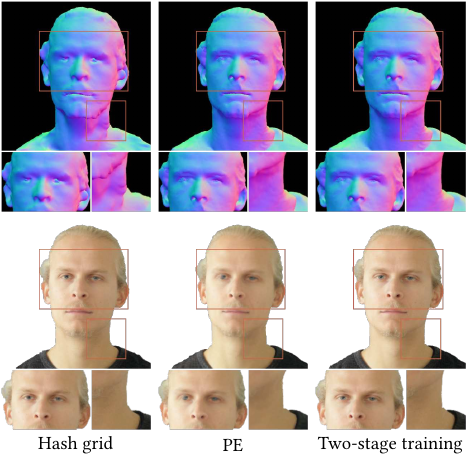}
\caption{Ablation Study. Qualitative comparison between the hash-grid encoding of \cite{mueller2022instant}, the positional encoding of \cite{Mildenhall_2020_nerf} (``PE''), and our two-stage approach. Top row: estimated normals; bottom row: estimated rendering.}
\label{fig:hash_pos_both}
\end{figure}
Through the course of our experiments, we noticed that it is necessary to control the speed at which the texture is learned in order to obtain both good geometry and albedo. In particular, using a hash-grid positional encoding \cite{mueller2022instant} results in better image quality, and the method converges very fast. However, this results in noisier geometries since there is not enough gradient signal coming from the color supervision. This behavior can be observed in the first column of Figure~\ref{fig:hash_pos_both}, where a high-quality rendered image corresponds to a relatively noisy geometry. On the other hand, using a standard positional encoding \cite{Mildenhall_2020_nerf} (second column in Figure~\ref{fig:hash_pos_both}) converges slower and leads to blurry textures, but learns geometric details from the observed images. Our two-stage training approach achieves the best of both options, as shown in the last column of Figure~\ref{fig:hash_pos_both}. 

\subsubsection{Mesh Upsampling}
\label{sec:meshres}
\begin{figure}
\centering
\newcommand{\mywidth}{0.23}
\includegraphics[page=1]{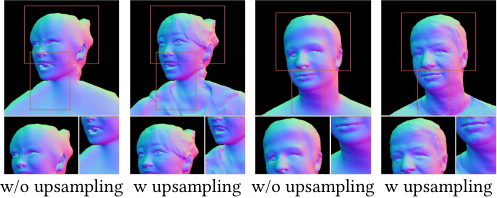}
\caption{Ablation Study: Mesh resolution. Qualitative comparison between surface normals of two subjects with and without upsampling the mesh. This figure demonstrates that upsampling the FLAME mesh to roughly 11K vertices helps capture high-fidelity geometry.}
\label{fig:ablation_wo_upsampling}
\end{figure}
The FLAME mesh contains 5023 vertices that model the face and neck region, without hair or shoulders. In our setting, we learn the geometry of the subjects including diverse hair types and hairstyles, facial hair, head accessories, and part of the shoulder. However, optimizing with only 5023 vertices results in a smooth coarse geometry, as illustrated in Figure~\ref{fig:ablation_wo_upsampling}. The output mesh appears smooth as the vertices around the shoulder and hair region are stretched out to form triangles occupying a large area. To capture the high-fidelity geometric details of the subject, we increase the resolution by upsampling the mesh \cite{Botsch2004ARA} to around 11k vertices. This improves the quality in the hair, neck, and shoulder regions. Note that our resolution is lower than the roughly 16K vertices used by NHA, yet our geometric quality is higher.
\begin{figure}
\centering
\newcommand{\mywidth}{0.23}
\includegraphics[page=1]{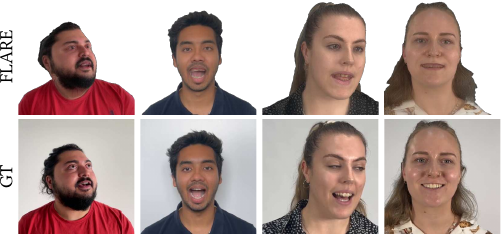}
\caption{Limitations. Modeling the mouth interior and eyes are challenging due to their complex material properties, variation in appearance (e.g.~subjects 1, 2, and 3 have different-sized teeth), and the fact that we do not model eye blinks (e.g.~subject 3). Capturing sharp specular highlights is also challenging due to the approximations made by our lighting model (subjects 3 and 4).}
\label{fig:limitations}
\end{figure}
\section{Limitations and Future Work}
\moniker{} can be trained in around 15 minutes and produces competitive results compared to methods that 
generate high-fidelity geometry at the expense of longer training times (on the order of days). However, there are still limitations, as shown in Figure~\ref{fig:limitations}. Firstly, the quality of the eyes and mouth interior needs improvement. These are challenging areas due to their complex material properties, and most neural avatar methods currently struggle with modeling these. For the mouth interior, an additional challenge comes from the fact that the teeth are exposed to varying degrees during training and this varies significantly between subjects. When a person does not smile with their teeth or does not articulate sufficiently, then the model does not have enough information to correctly reproduce the tooth color and geometry. Further, the FLAME mesh does not have vertices in the mouth interior and thus, during rasterization, there are no vertices projected onto the image of the mouth, resulting in no gradient being propagated there. Our remeshing step partly addresses this problem and, for some subjects, there are vertices formed around the teeth. However, modeling the teeth remains a challenging task due to the constant motion of the lips and limited supervision. Similarly, the eye area exhibits challenging photometric properties that are not always captured by our method. In addition, our pre-processing step does not track eye blinking, resulting in inevitable errors during optimization that yield a noisy geometry around the eyes. Future work should develop techniques that can enhance the estimation of the mouth and eye area, in both photometric and geometric respects. 

Second, capturing harsh neck shadows, self-shadows, and sharp specular highlights is difficult as demonstrated in Figure~\ref{fig:limitations}. We can remove shadows cast on the face region as the subject moves their head in various directions. However, the shoulder and neck areas remain mostly static and shadows are baked in. Additionally, although non-Lambertian reflections that look plausible can be captured by our method during relighting due to the estimated materials,  we miss reproducing the sharp specular highlights of the ground truth. This is due to the several approximations that we make to model the pre-filtering of the environment and to simplify the integral of the rendering equation. 
Finally, our method does not model more subtle skin properties such as sub-surface scattering, or time-dependent appearance changes. We hypothesize that this could enhance realism and, consequently, the estimated geometry. We believe this is an interesting direction to pursue in the future. 
\section{Ethics}
The goal of \moniker{} is to enable fast, subject-specific, avatar creation that can be used to generate novel expressions and to place the avatars in different scenes. 
This capability, however, opens the door to potential misuse, where new malicious content of the training subject can be generated without their consent. 
Although the quality of \moniker{} still exhibits identifiable artifacts signaling its AI origin,  the rapid progression of the field suggests these cues may diminish over time. Addressing this remains an important technical and legal challenge. 
\section{Conclusion}
In this work we presented \moniker{}, a new method for building animatable and relightable head avatars from monocular video in $15$ minutes. Our approach directly produces a mesh representation that can be efficiently rendered and animated, along with material parameters that allow the  avatars to be placed in scenes under novel illumination. This is achieved by combining traditional computer graphics methods for rendering with neural networks that approximate some of the components. More specifically, 
we optimize a canonical mesh geometry while approximating the expression deformations, albedo, roughness and specular intensity values using coordinate-based MLPs. Further, we avoid explicitly computing an environment map from a narrow field of view by approximating the pre-filtered environment map in the split-sum formulation with a neural network. Finally, we propose a two-stage approach designed to control the pace at which geometry and texture are learned relative to one another.
Our experimental results show that we can obtain mesh avatars of high geometric and image fidelity. Once learned, the avatars can be readily inserted and rendered in arbitrary scenes using standard graphics pipelines, enabling downstream applications in gaming, film production and telepresence. 

\begin{acks}
We thank Jacob Munkberg, Jon Hasselgren, and Pramod Rao for fruitful discussions and Wojciech Zielonka for  discussions regarding the baseline. We thank Asuka Bertler, Claudia Gallatz, Taylor McConnell, and Markus Höschle for their support with data collection and thank all the participants for their time. We thank Haoran Yun, Peter Kultis, and Nikos Athanasiou for additional support. 
\paragraph{Disclosure:} MJB has received research gift funds from Adobe, Intel, Nvidia, Meta/Facebook, and Amazon.  MJB has financial interests in Amazon, Datagen Technologies, and Meshcapade GmbH.  While MJB is a consultant for Meshcapade, his research in this project was performed solely at, and funded solely by, the Max Planck Society.
\end{acks}
\bibliographystyle{ACM-Reference-Format}
\bibliography{mybibliography}
\appendix
\balance
\section{Implementation Details}
\paragraph{FLAME Mesh and Deformation Network.}
 We manually add mesh faces to the FLAME template mesh between the upper and lower lips to close the mouth cavity, similar to NHA~\cite{Grassal_2022_nha}. Additionally, we also simplify the tessellated eye region of the FLAME template following \cite{zielonka2022insta}.  Similar to PointAvatar \cite{zheng2023pointavatar}, during training, we map the optimized mesh vertices to a canonical pose with jaw open and a neutral expression and then proceed to perform LBS to deform the mesh. This additional step encourages the canonical mesh to have an open-mouth expression, which facilitates the learning of mouth movements. 
 We train the deformation MLP $\dmlp$ only during the first stage with a learning rate of $10^{-3}$ and use the Adam optimizer \cite{kingma14}. We adopt the network architecture of PointAvatar which is similar to \cite{Zheng_2022_imavatar}, except we do not predict additional vertex displacements (only skinning weights, expression and pose blendshapes). During the second stage, we freeze the deformation network and use the weights from the first stage. 

 \paragraph{Optimization of Mesh Vertices.}
The canonical mesh is upsampled once during training, resulting in a final mesh of approximately 11K vertices. During the first stage of training, when the number of vertices increases, we reduce the learning rate of the vertex offsets from $10^{-3}$ to $10^{-3} * 0.75$ and increase the weight of the Laplacian and normal regularizer by 4 times following \cite{worchel2022_nds}. This helps in learning a smoother mesh and prevents the vertices from diverging after the upsampling step. During the second stage, we set the learning rate of the vertex offsets to a very small value ($10^{-5}$) and initialize the training with the canonical mesh from the previous stage. 

\paragraph{Texture Estimation.}
For the Material MLP $\fmlp$ during the first stage, we use a ReLU MLP of 4 hidden layers with 128 neurons each. For the final layer, we use the sigmoid activation function. For the second stage, since we use the hash-grid positional encoding, we adopt a smaller network architecture of 2 hidden layers of 64 neurons each with the same activation functions as before. We set the learning rate at $10^{-3}$ and use the Adam optimizer. The Fresnel coefficient is set to $F_0 = 0.047$ during the first stage following \cite{karis2013real}. During the second stage, since the geometry is well-estimated and the rendered shape aligns well with the ground truth, we use a segmentation mask and explicitly set the Fresnel coefficient of the skin region to $F_0 = 0.028$, while the rest is set to a constant of $F_0 = 0.047$ (Fresnel coefficient for hair). For the lighting MLP $\smlp$, we use a ReLU MLP with 2 hidden layers with 64 neurons each for both training stages. However, we do not use an activation function for the output layer as we learn these computations in the sRGB log space of tone-mapped RGB colors.

\section{Data Pre-processing}
\label{ap:preprocessing}
We use the pre-processing pipeline of IMavatar \cite{Zheng_2022_imavatar} where the segmentation masks are obtained from MODNet \cite{MODNet}, the FLAME parameters (shape, pose and expression) and the camera parameters are estimated using DECA \cite{feng2021learning} and later refined by fitting to 2D facial keypoints \cite{Bulat_2017}. For IMavatar, NHA, PointAvatar, and our method, we use the same pre-processed data. We refrain from using the same pre-processed data for INSTA \cite{zielonka2022insta} because the method does not optimize for pose parameters and instead translates the camera. Moreover, it has a depth-supervision loss where the depth maps are generated from the shape parameter of FLAME, and the preprocessing used by INSTA (MICA: \cite{MICA:ECCV2022}) is quantitatively better than the shape estimation of DECA \cite{feng2021learning}. Thus, to get the best results for INSTA, we use the pre-processing released by the authors. Additionally, the synthetic dataset \cite{makehuman} contains dynamic neck motions, and INSTA's pre-processing fails to estimate the camera pose correctly for many frames. Hence, for INSTA, we only evaluate the frames that have a mask overlap of more than 90\%. 
\end{document}